\documentclass[conference]{IEEEtran}

\IEEEoverridecommandlockouts                              

\usepackage{multirow}
\usepackage{graphicx}
\usepackage{subcaption}
\usepackage{svg}

\usepackage{xcolor}

\definecolor{mygreen}{RGB}{0, 180, 0}

\newcommand{\fengyi}[1]{{\color{black} #1}}
\newcommand{\gordonmod}[1]{{\color{black} #1}}

\title{\LARGE \bf
	A Bio-mimetic Neuromorphic Model for Heat-evoked Nociceptive Withdrawal Reflex in Upper Limb}

\author{Fengyi Wang$^{1}$, J. Rogelio Guadarrama Olvera$^{1}$\gordonmod{, Nitish Thakor$^{2}$ and Gordon Cheng$^{1}$ }
	\thanks{$^{1}$Fengyi Wang,  J. Rogelio Guadarrama Olvera, and Gordon Cheng are with the Institute for Cognitive Systems, Technical University of Munich, Arcisstrae 21, 80333 Munich, Germany
		{\tt\small \{ fengyi.wang, rogelio.guadarrama, gordon \} @tum.de}
	} 
}

\begin{document}
	\maketitle
	\thispagestyle{empty}
	\pagestyle{empty}

	\begin{abstract}
		The nociceptive withdrawal reflex (NWR) is a mechanism to mediate interactions and protect the body from damage in a potentially dangerous environment. To better convey warning signals to users of prosthetic arms or autonomous robots and protect them by triggering a proper NWR, it is useful to use a biological representation of temperature information for fast and effective processing. In this work, we present a neuromorphic spiking network for heat-evoked NWR by mimicking the structure and encoding scheme of the reflex arc. The network is trained with the bio-plausible reward modulated spike timing-dependent plasticity learning algorithm. We evaluated the proposed model and three other methods in recent studies that trigger NWR in an experiment with radiant heat. We found that only the neuromorphic model exhibits the spatial summation (SS) effect and temporal summation (TS) effect similar to humans and can encode the reflex strength matching the intensity of the stimulus in the relative spike latency online. The improved bio-plausibility of this neuromorphic model could improve sensory feedback in neural prostheses.
		
	\end{abstract}

	\section{Introduction}

	Far beyond the initial developments, modern prostheses are beginning to allow those with upper limb amputations to regain basic functionality through intuitive control \cite{cordella_literature_2016}. The recently developed prosthesis could even provide perceptual touch feedback and protection against sharp objects with a withdrawal reflex \cite{osborn_prosthesis_2018}. However, preventing damage caused by heat has received less attention.
	
	The nociceptive withdrawal reflex (NWR) is a polysynaptic spinal reflex that withdraws the affected body part away from the stimulation site \cite{shahani1971human}. The external thermal stimulations are encoded into electrical activities by the heat-sensitive neurons in the skin, including mainly three groups of heat-sensitive neurons that respond to different temperatures. As temperature increases, more heat-sensitive neurons are activated, and most individual neurons respond more strongly \cite{wang_sensory_2018}. The interneuron in the spinal cord collects the sensory signals and projects the encoded reflex strength to the motor neurons \cite{schouenborg_somatosensory_2003}. The evoke of NWR is not simply gated by a fixed temperature threshold but a consequence of the excitability of spinal reflex arc, descending control, and the dynamic of the sensory input \cite{andersen_studies_2007}. Furthermore, the reflex strength is related to the stimulation intensity \cite{campbell_comparison_1991}.

	Several methods of detecting noxious stimuli and triggering reflexes have been implemented in prosthetic or robotic systems to avoid unintended damage. \cite{junge_bio-inspired_nodate} uses analog circuits that bypass the central processor to achieve fast reflex. \cite{neto_skininspired_2022} distinguish between three temperature ranges with a neural classifier. \cite{osborn_prosthesis_2018} mimics the reflex arc with spiking neurons to protect the prosthetic hand against sharp objects.
	
\begin{figure}
		\centering
		\includegraphics[width=0.7\linewidth]{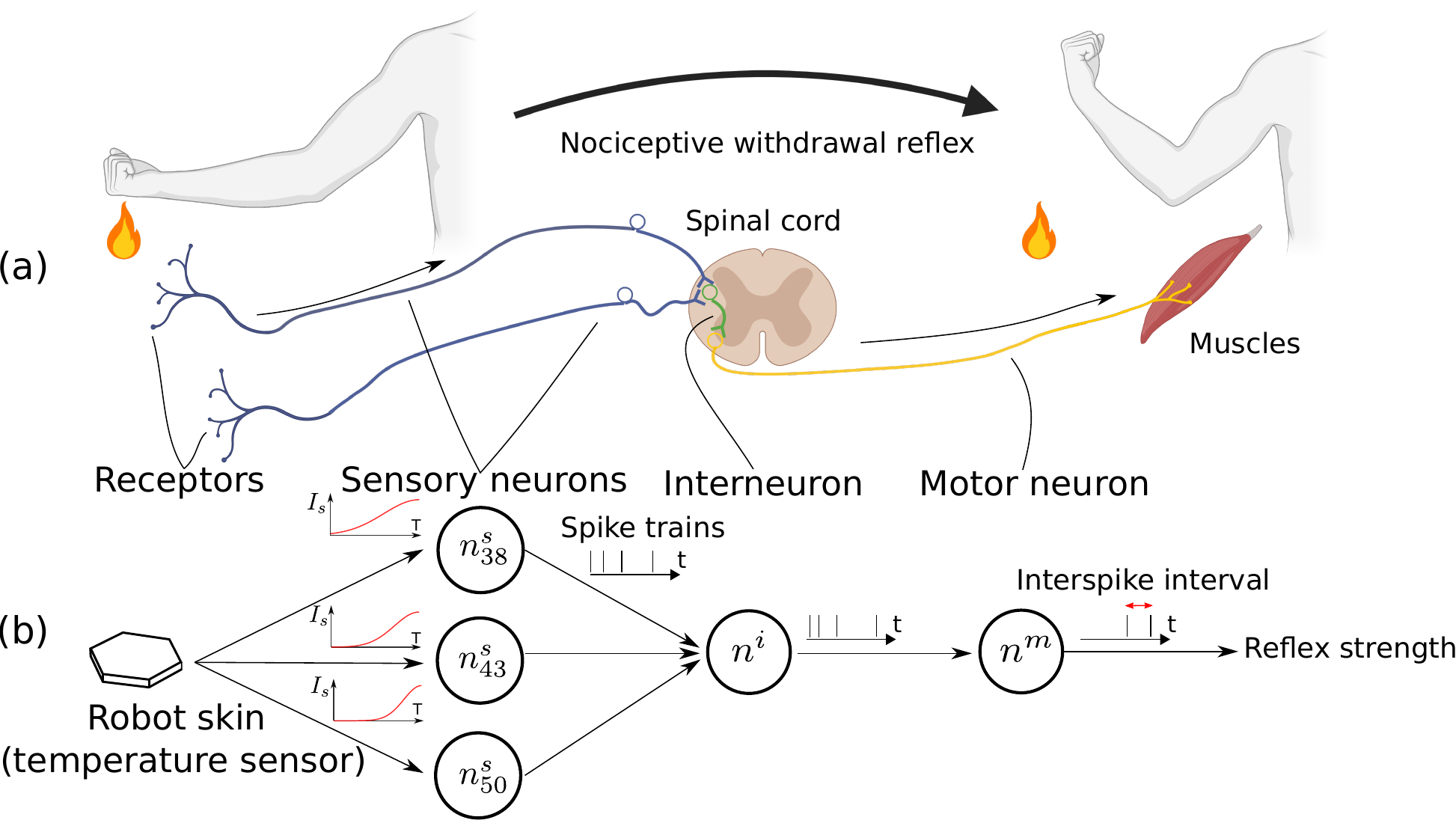}
		\caption{Illustration of the polysynaptic reflex arc and the proposed bio-mimetic neuromorphic model of NWR.}
		\label{fig:network}
	\end{figure}

	In this paper, we introduce a neuromorphic model that mimics the firing pattern of individual neurons and the structure of the reflex arc as illustrated in Fig. \ref{fig:network}. The network is trained with the bio-plausible reward-modulated spike timing dependent plasticity (R-STDP) algorithm to generate the reflex strength depending on the intensity of external stimulation. The proposed method utilizes the intrinsic characteristics of the spiking neural network to reproduce several essential features of the NWR in humans without additional modeling or training. We demonstrate the benefits of using the neuromorphic model over three other reflex-triggering methods in recent studies. Our work has implications for prosthetic limbs as well as humanoid robotic systems that utilize thermal information to avoid potential damage.
	
	\section{Method}
	\subsection{Bio-mimetic spiking neural networks}

	\subsubsection{Sensory neurons} \label{subsubsection:sensoryneurons}
	
	Biological heat-sensitive neurons have a burst-adaption firing pattern to changing temperature \cite{konietzny_dynamic_1977}, so we use the Izhikevich neuron model with the intrinsically bursting spiking pattern \cite{izhikevich_simple_2003} to convert the measured temperature into neuromorphic signals.
	
	The most heat-sensitive neurons respond to are 38$^\circ$C, 43$^\circ$C and 50$^\circ$C, respectively \cite{wang_sensory_2018}. We model these neurons that respond to different temperatures by assigning them different tuning curves with the half-Gaussian-like function (\ref{eq:tuning1}). For a sensory neuron $n_{T_a}^s$ that responds to temperature $T_a$, the virtual input current is:
	
	\begin{equation}
	\label{eq:tuning1}
	I^{s}_{T_a}=\exp \left(-\frac{1}{2} \frac{(T_{max} - T)^{2}}{\sigma^{2}}\right) \cdot I^{s}_{max} 
	\end{equation} 
	\label{eq:tuning2}
	$\textrm{if\ } T \geq T_{max},\  \textrm{then}\  I^{s}_i = I^{s}_{max}$.
	
	In (\ref{eq:tuning1}), $T$ is the input temperature, $T_{max}$ is set to $52^\circ$C  according to the activation threshold of the TRPV2 ion channel, which is related to the detection of high temperature \cite{basbaum_cellular_2009}. $I^{s}_{T_a}$ represents the input current of the sensory neuron $n_{T_a}^{s}$ which responded to $T_a$. For $n_{T_a}^{s}$, $\sigma_i = T_{max} - T_a$, so that all neurons have the same input current at $T_a$.
	
	\subsubsection{Interneuron}
	Interneurons in primates also have a burst-adaption firing pattern \cite{takei_neural_2017}. Thus, we use the same mathematical model to describe the internal states of the interneuron in the proposed model.
	
	The sensory information encoded in the spike trains and the descending control converges in the interneuron. The input current of an interneuron is increased when it receives a spike from a sensory neuron. Otherwise, the input current decreases over time. 
	
	The amplitude of the post-synaptic current change can be determined by the synaptic strengths $s_{T_a}$ between the sensory neuron that responded to ${T_a}$ and the interneuron. We use a parameter $\epsilon$ to represent the overall effect of the descending control and the excitability of the spinal reflex arc. While $\epsilon=1$ is the normal state, $\epsilon>1$ means that the NWR is facilitated, and $\epsilon<1$ means that the NWR is inhabited. The input current of the interneuron $I^i$ is described by the following equation:
	\begin{equation}
	d I^i/d t=- I^i/\tau_i
	\end{equation}
	$	\textrm{when\ } n^s_{T_a} \textrm{\ spikes,\ }I^i = I^i + \epsilon \cdot s_{T_a} \cdot I_{max} $
	
	\subsubsection{Motor neuron}
	
	This layer contains a leaky integrate and fire (LIF) neuron $n^m$ that mimics an $\alpha$ motor neuron that evokes the NWR in humans \cite{serrao_kinematic_2006}. The dynamic of the motor neurons can be described with the following equations:
	\begin{equation}
	d v_m/d t=(v_0 - v)/\tau_m
	\end{equation}
	$\textrm{if\ } v_m \geq v_{th}^m,\  \textrm{then}\  v_m = v_0$
	
	where the resting potential $v_0$ = -70 mV and the time constant ${\tau_m}$ = 50 ms which is presented in \cite{izhikevich_solving_2007}. A spike from the interneuron induces a 30 mV increment in the membrane potential of the motor neuron $v_m$. When $v_m$ reaches the threshold $v_{th}^m$, the motor neuron generates a spike, and the membrane potential $v_m$ is reset to the resting potential $v_0$. Fig. \ref{fig:network} (b) shows the overall structure of the spiking neural network.
	
	\subsection{Learning algorithm} \label{subsection:learning algorithm}
	
	In an STDP synapse, the modification of synaptic strength $s_{i}$ that arises from a pair of pre- and post-synaptic spikes with the inter-spike interval $\Delta t$ can be expressed by the function $F(\Delta t)$ \cite{song2000competitive}:

	\begin{equation}
	F(\Delta t)=\left\{\begin{array}{l}
	\mathrm{A}_{+} \exp \left(\Delta t / \tau_{+}\right) \textrm {if } \Delta \mathrm{t}<0 \\
	\mathrm{A}_{-} \exp \left(-\Delta t / \tau_{-}\right) \textrm {if } \Delta \mathrm{t} \geq 0
	\end{array}\right.
	\end{equation}
	
	where $\mathrm{A}_{+}$, $\mathrm{A}_{-}$ determine the maximum amount of synaptic modification and $\tau_{+}$, $\tau_{-}$ determine the range of the pre-to-post inter-spike interval over which the synaptic connection is modified.
	
	The reward modulation of an STDP synapse can be described by  the following differential equations:
	
	\begin{equation}
	\left\{\begin{array}{l}
	d C/d t=- C/\tau_C \\
	d e/d t=- e/\tau_e \\
	d s/d t=C\cdot e /\tau_s
	\end{array}\right.
	\end{equation}
	
	where $C$ represents the concentration of extracellular dopamine, i.e., the intensity of the reward/punishment, and $e$ represents the eligibility trace for synaptic modification \cite{houk199513}. The decay rates $\tau_C$, $\tau_e$, and $\tau_s$ control the sensitivity of the plasticity to the delayed reward/punishment. 
	
	The reflex strength $\rho$ increase exponentially as a function of the intensity of the stimulation in human:
	$\rho=\mathbf{\alpha}\cdot \beta^{(T-T_{0})}$
	\cite{campbell_comparison_1991}. where $T$ is the stimulation intensity and $T_{0}$ is the threshold of pain \cite{stevens1961honor}. We define normalized reflex strength $|\rho|$ as the ratio of the reflex strength to the reflex strength at $T_{max}$:
	\begin{equation}
	|\rho| = \frac{\rho}{\rho_{max}} = \frac{\mathbf{\alpha}\cdot \beta^{(T-T_{0})}}{\mathbf{\alpha}\cdot \beta^{(T_{max}-T_{0})}} =\beta^{(T-T_{max})}
	\end{equation}
	
	The normalized reflex strength $|\rho|$ is independent of $\mathbf{\alpha}$ and threshold $T_0$. The exponent $\mathbf{\beta}$ is set to $1.268$ as in \cite{campbell_comparison_1991}.
	
	In the learning phase, the motor neuron does not fire so that a continuous membrane potential $v_m$ can be obtained. $v_m$ is normalized by a scaling factor $K_m$ to match the scale of $|\psi|$ and the threshold of the motor neuron:
	\begin{equation}
	|v_m| = (v_m + v_{th}^m)/K_m
	\end{equation}
	
	We define an ideal stimulation with magnitude $T_s$ applied on time $t_0$ and last for $t_s$ is defined as:
	\begin{equation}\label{eq:idealstimulation}
	T = T_s \cdot (H(t-t_0) - H(t-t_0-t_s))
	\end{equation}
	where $H(t)$ is the Heaviside step function.
	
	Ideal stimulations of random intensities that lasts for $100$ ms followed by a cooling phase that lasts for $2$ seconds at $35 ^\circ$C are presented to the network repetitively. The normalized maximum value of the membrane potential $|{v_m}|_{max}$ during the stimulation phase is defined as the normalized activation level of the motor neuron.
	
	After each stimulation phase, a reward/punishment signal is given. The intensity of the reward/punishment is determined by the relationship between the normalized reflex strength $|\rho|$ and the activation level $|{v_m}|_{max}$. 
	\begin{equation}
	C =|\rho| - |{v_m}|_{max} 
	\end{equation}
	
	\section{Experimental evaluation}
	
	In the experiments with realistic heat stimulation, the temperature is sensed by a robot skin cell \cite{cheng_comprehensive_2019} and conveyed to the sensory neurons at 100 Hz. A lamp of 1kW power is used to heat up the robot skin cell. The neuromorphic model is implemented in C++ using the Robot Operating System (ROS).
	
	\subsection{Reflex strength}
	
	To demonstrate that the proposed model can generate the reflex strength matching the stimulation intensity, we apply multiple ideal stimulations (\ref{eq:idealstimulation}) of different intensities. The strength of the NWR is determined by the reflex strength $g_s$ decoded from the relative spike latency $\Delta t$ between spikes of the motor neuron:
	\begin{equation}
	g_s = \cdot \beta^{(\Delta t_{min} - \Delta t)/k}
	\end{equation}
	
\begin{figure}
		\centering
		\includegraphics[width=0.75\linewidth]{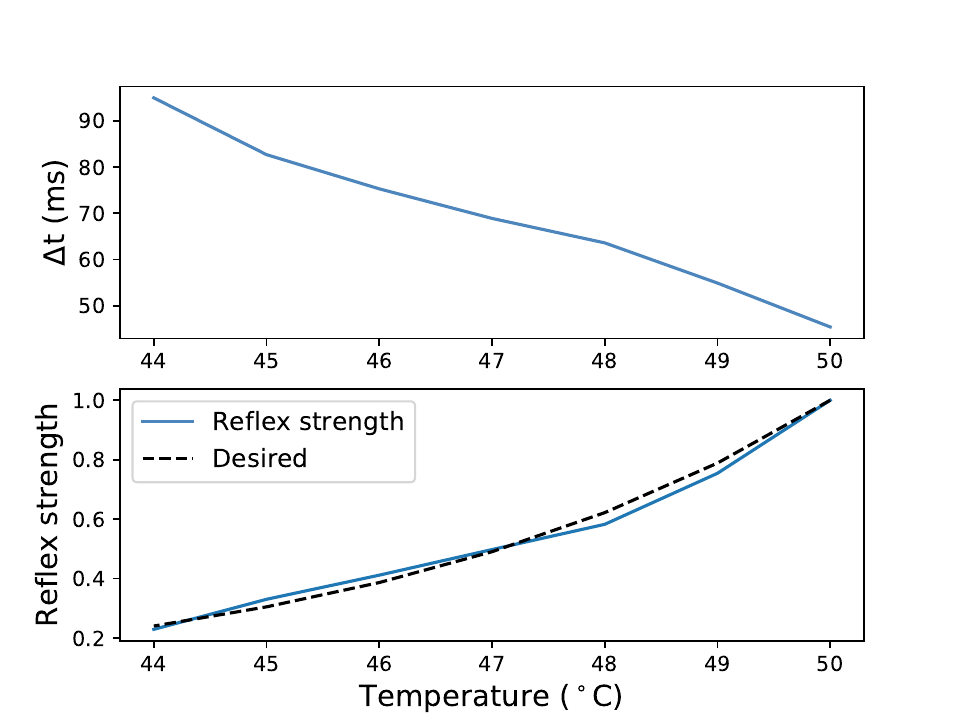}
		\caption{Lines in (a) and (b) shows how $\Delta t$ and reflex strength $g_s$ varies with stimulation intensity, respectively. The dashed line in (b) shows the NWR strength in the human forearm \cite{campbell_comparison_1991}.
		}
		\label{fig:strength}
	\end{figure}

	The result is shown in Fig. \ref{fig:strength}, the relative spike latency $\Delta t$ decreases as the stimulation intensity increases. Consequently, the reflex strength $g_s$ increases approximately exponentially, which is consistent with the strength of NWR evoked by radiant heat in the forearm \cite{campbell_comparison_1991}.
	
	\subsection{Spatial summation effect}
	
	The phenomenon that more neurons are activated under high-intensity stimulation and thus facilitate reflexes is called the spatial summation (SS) effect \cite{andersen_studies_2007}.
	
	To demonstrate the SS effect under high temperature, $\epsilon$ is set to 0.8. The experimental results in Fig. \ref{fig:result}$(b)$ shows that the sensory neuron with $T_a = 50^{\circ}$C (neuron index 3) is recruited under high temperature to evoke the NWR under high temperature.

	\subsection{Temporal summation effect}
\begin{figure}
		\centering
		\includegraphics[width=0.85\linewidth]{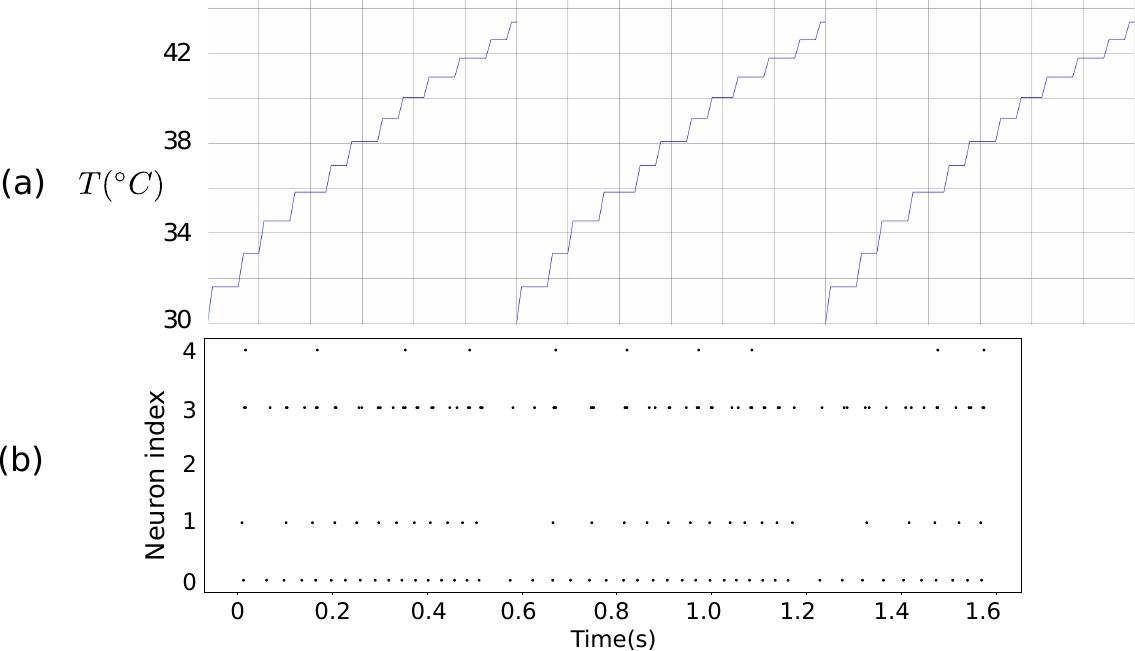}
		\caption{ (a) and (b) illustrate the stimulation in the SS effect experiment and the corresponding output of the neurons in the model.}
		\label{fig:ts}
	\end{figure}

	Besides the SS effect, NWR in humans can be evoked by repeated sub-threshold stimulation, which is called the temporal summation (TS) effect \cite{arendt2000facilitation}.
	In this TS experiment, part of the stimulation recorded in the SS effect experiment, whose peak value is below the reflex threshold, is sent to the model. The stimulation is illustrated in Fig. \ref{fig:ts}(a). 
	
	We found that such repeated sub-threshold stimulation can evoke an NWR. \fengyi{The decrease in sensory neuron input current is comparable with the relationship between threshold and frequency in \cite{arendt2000facilitation}.} The applied stimulation and the raster of the proposed neural network are shown in Fig.\ref{fig:ts}. This effect shows that the model is more sensitive to rapidly changing temperatures, helping to detect very hot objects with high temperature gradients. 
	
	\subsection{Comparison}
	
	In this part, we compare the proposed method with three other methods that trigger nociceptive reflex with the same thermal stimulation recorded in a ROS bag.

	\subsubsection{Analog reflex system}
	
\begin{figure}
		\centering
		\includegraphics[width=0.7\linewidth]{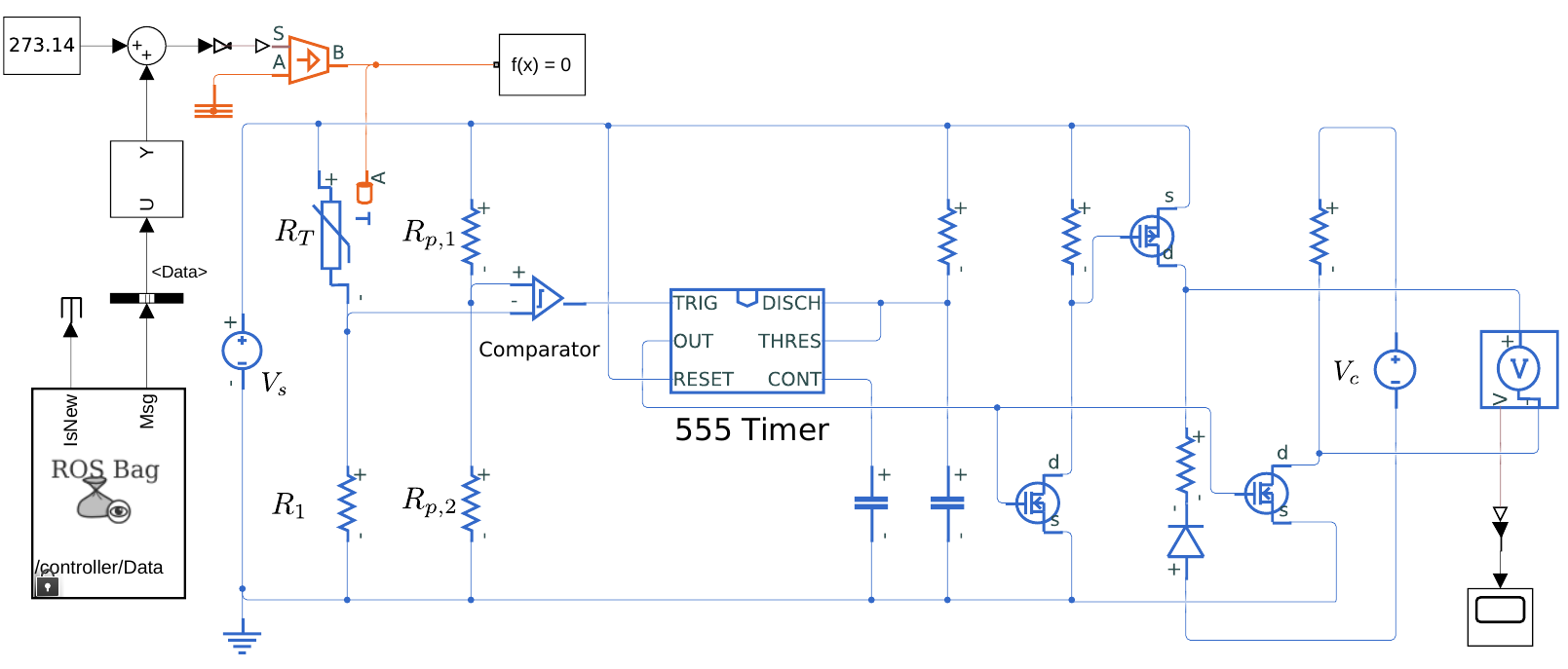}
		\caption{The simulation of the analog reflex circuit. $V_S$ represents the voltage applied to the motor during the reflex action. $V_c$ represents the motor command from the microcontroller.}
		\label{fig:analog}
	\end{figure}

	\cite{junge_bio-inspired_nodate} uses an analog reflex system to minimize the damage to the robotic gripper when it directly interacts with hot objects. We reproduce this implementation using MATLAB/Simulink, as shown in Fig. \ref{fig:analog}.

	\begin{figure}[h]
		\centering
		\includegraphics[width=0.7\linewidth]{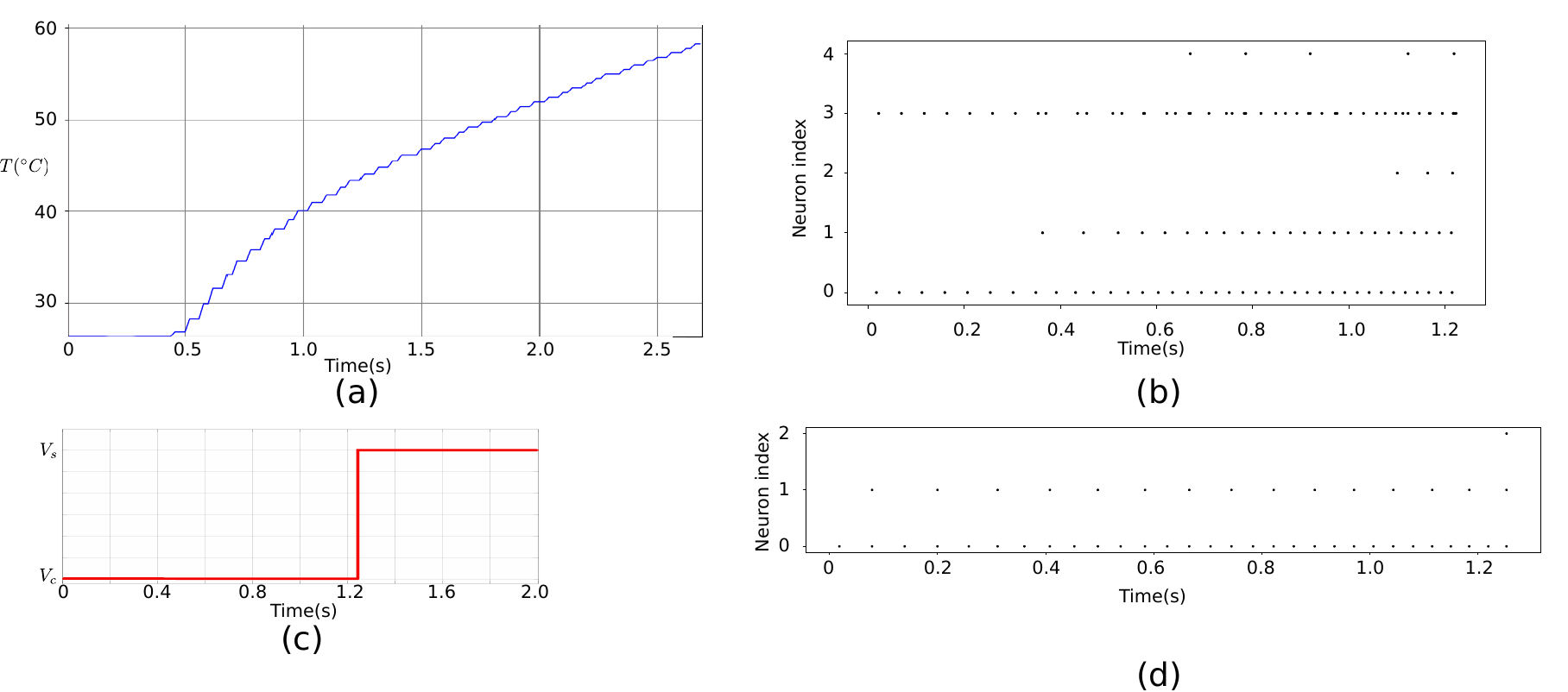}
				\caption{(a), (b), (c), and (d) illustrate the stimulation and the output of the proposed neuromorphic model, analog system, and feed-forward spiking neurons, respectively.}
		\label{fig:result}
	\end{figure}
	In this system, the gripper driven by a DC motor is controlled by a microcontroller (Arduino Nano), and a thermistor is used as a sensory receptor. The resistance of a negative temperature coefficient thermistor can be characterized by the following:
	\begin{equation}
	R_T=R_0 e^{B\left(1 / T-1 / T_0\right)}
	\end{equation}
	where $B$ is a parameter and $R_0$ is the resistance at $T_0$.
	
	When the voltage on $R_1$ exceeds the reference voltage set by a potentiometer (represented by $R_{p,1}$ and $R_{p,2}$), the comparator generates a falling edge and resets the latch of the 555 timer in the circuit. The motor command signal from the microcontroller is overridden, and the reflex action is triggered. Thus, the threshold of the reflex action is:
	\begin{equation}
	T_{thres}=B\cdot T_0/(B + T_0 \cdot ln(R_1\cdot R_{p,1}/R_0\cdot R_{p,2}))
	\end{equation}
	
	We set $R_0 = 1\ k\Omega$, $T_0 = 298.15\ K$, $B = 3500\ K$, $R_1 = 500\ \Omega$ and $R_{p,1} = R_{p,2} = 10\ k\Omega$. The threshold in the simulation is 43.7 $^\circ$C, which is consistent with the expected value. The result is illustrated in Fig. \ref{fig:result}(c).
	
	\subsubsection{Neural classifier}
	\cite{neto_skininspired_2022} uses a neural classifier containing two perceptrons driven by the same input to distinguish between 3 temperature ranges (cold, room temperature, and hot). Each perceptron is a binary classifier trained with the back-propagation algorithm. A classification function (\ref{eq:ensemble}) is used to ensemble the output from two perceptrons (N1 and N2).
	\begin{equation}
	\label{eq:ensemble}
	f(N 1, N 2)= N 1+N 2
	\end{equation}
	
	We train a classier with the given training data and parameters. We find that the network is overfitted during the training, \fengyi{and the classifier has fixed thresholds of cold and hot}, which are $22.5\ ^\circ$C and $107/3 \ ^\circ$C, respectively.
	
	\subsubsection{Feed-forward spiking neurons}
	\cite{osborn_prosthesis_2018} uses feed-forward spiking neurons to distinguish between a noxious and innocuous stimulation. We reproduced the model and successfully tuned the threshold to be 46.8 $^\circ$C. The experiment result is shown in Fig. \ref{fig:result}(d). 
	
	However, the lack of a training algorithm makes the model relies on manually tuned parameters. The simple structure only allows it to make a binary classification. This model is less sensitive to the dynamics of the stimulation because of the feature of the LIF model, and we did not find significant temporal summation effects in this model.

	\section{Conclusion and discussion}
	
	This paper introduces a bio-mimetic neuromorphic model of heat-evoked withdrawal reflex trained with the bio-plausible R-STDP algorithm and converts the input signal from sensors into biologically relevant signals. This paper also compares the proposed model and three reflex-triggering methods in recent studies by testing them with the same stimulation in the experiments. The experiment shows that the proposed model reproduces the spatial and temporal summation effects of NWR in humans. The relationship between threshold and frequency is comparable with that in humans. While all the comparing methods make only binary classifications with fixed thresholds, and the proposed neuromorphic mode generates reflex strength according to the intensity of the stimulation. This improvement is largely due to the bio-mimicking population encoding scheme. Furthermore, the neuromorphic implementation that utilizes the temporal information encoded in the spike trains makes it possible to process multi-modal information on low-power neuromorphic platforms. We envision that this model will be a preliminary study for the development of neuromorphic algorithms for autonomous robotic systems and human-machine cooperation applications.
	
	\section{Acknowledgement}
	This work was supported by the German Federal Ministry of Education and Research (BMBF) under Grant 01GQ2108.
	
	
	
	\bibliographystyle{ieeetr}
	\bibliography{ref}
	
\end{document}